\documentclass[]{interact}

\usepackage{epstopdf}% To incorporate .eps illustrations using PDFLaTeX, etc.
\usepackage[caption=false]{subfig}% Support for small, `sub' figures and tables
\usepackage{multicol}

\usepackage{algorithmicx}
\usepackage{algorithm,algpseudocode}
\usepackage[export]{adjustbox}
\usepackage{xcolor}

\usepackage{natbib}% Citation support using natbib.sty
\bibpunct[, ]{(}{)}{;}{a}{}{,}% Citation support using natbib.sty
\renewcommand\bibfont{\fontsize{10}{12}\selectfont}% Bibliography support using natbib.sty

\theoremstyle{plain}% Theorem-like structures provided by amsthm.sty

\theoremstyle{definition}

\theoremstyle{remark}

\usepackage{hyperref}

\algnewcommand\algorithmicforeach{\textbf{for each}}
\algdef{S}[FOR]{ForEach}[1]{\algorithmicforeach\ #1\ \algorithmicdo}

\newcommand{\symfootnote}[1]{%
\let\oldthefootnote=\thefootnote%
\stepcounter{mpfootnote}%
\addtocounter{footnote}{-1}%
\renewcommand{\thefootnote}{\fnsymbol{mpfootnote}}%
\footnote{#1}%
\let\thefootnote=\oldthefootnote%
}

\begin{document}

% \articletype{ARTICLE TEMPLATE}% Specify the article type or omit as appropriate

% \title{AMBF+: Surgical Simulation Framework with Computer Vision Data Generation}
\title{Virtual Reality for Synergistic Surgical Training and Data Generation}

\author{
\name{Adnan Munawar\textsuperscript{a}\thanks{ Email: \{amunawar,\,zli122\}@jhu.edu}, Zhaoshuo Li\textsuperscript{a}, Punit Kunjam\textsuperscript{a}, Nimesh Nagururu\textsuperscript{a}, Andy S. Ding\textsuperscript{a}, Peter Kazanzides\textsuperscript{a}, Thomas Looi\textsuperscript{b}, Francis X. Creighton\textsuperscript{a}, Russell H. Taylor\textsuperscript{a} and Mathias Unberath\textsuperscript{a}}
\affil{\textsuperscript{a}Johns Hopkins University, Baltimore, US; \\ \textsuperscript{b}Hospital for Sick Children, Toronto, Canada.}
}

\maketitle

\begin{abstract}
Surgical simulators not only allow planning and training of complex procedures, but also offer the ability to generate structured data for algorithm development, which may be applied in image-guided computer assisted interventions. While there have been efforts on either developing training platforms for surgeons or data generation engines, these two features, to our knowledge, have not been offered together. We present our developments of a cost-effective and synergistic framework, named Asynchronous Multibody Framework Plus (AMBF+), which generates data for downstream algorithm development simultaneously with users practicing their surgical skills. AMBF+ offers stereoscopic display on a virtual reality (VR) device and haptic feedback for immersive surgical simulation. It can also generate diverse data such as object poses and segmentation maps. AMBF+ is designed with a flexible plugin setup which allows for unobtrusive extension for simulation of different surgical procedures. We show one use case of AMBF+ as a virtual drilling simulator for lateral skull-base surgery, where users can actively modify the patient anatomy using a virtual surgical drill. We further demonstrate how the data generated can be used for validating and training downstream computer vision algorithms.\symfootnote{Code is available at \url{https://github.com/LCSR-SICKKIDS/volumetric_drilling}}
\end{abstract}


\section{Introduction}
Effective ``training'' is pertinent for both surgeons and image-guided intervention algorithms. Thus, creating opportunities to learn from experience for both surgeons and algorithms has become of substantial interest. 

On one hand, surgeons can develop necessary surgical skills and 3D spatial perception by using surgical simulators with the benefit of safety of a simulated world (\cite{chan2016high}, \cite{lui2017evaluating}, \cite{locketz2017anatomy}, \cite{chen2018patient}). Beyond improving general surgical skills, surgical simulators can improve outcomes in individual patients through the ability to generate surgical views from unique pre-operative scans. More recently, Asynchronous Multi-Body Framework (AMBF) (\cite{munawar2019real}), has recently been introduced to the field and has been designed for fluid interaction with virtual surgical phantoms.

On the other hand, developing algorithms for image-guided computer-assisted surgical interventions also requires ``training''. It generally requires structured data either for validation purposes used in geometric-based algorithms such as Simultaneous Localization and Mapping (SLAM); or for training supervision used in deep-learning based algorithms, such as learning based segmentation (\cite{zisimopoulos2017can}). Many prior work in synthetic medical data generation exist for different modalities, such as CT scans (\cite{unberath2018deepdrr}) and pathology slices (\cite{mahmood2019deep}) as summarized in \cite{chen2021synthetic}. VisionBlender by \cite{cartucho2020visionblender} uses Blender\footnote{https://www.blender.org/} to generate data for endoscopic surgeries, \textit{off-line}, for development of computer vision algorithms. The images generated are visually pleasing however motion of camera and surgical instruments is only possible along pre-specified paths. Realistic tool-tissue interaction and camera motion enable cost-effective development or analysis for algorithms contingent on these information. For example, SurgVisDom\footnote{https://surgvisdom.grand-challenge.org/} is a synthetic dataset with realistic tool motion designed for surgical activities recognition, yet only designed for robotic laparoscopic surgery and not open-sourced.

Integrating the two components, \textit{i.e.} generating data from realistic surgical simulation with tool-tissue interaction in real-time, will offer a synergistic, and therefore cost-effective solution to both problems. While surgeons practice their surgical skills using the simulator, clinically relevant data are collected in a precisely controlled environment for downstream development of computer vision algorithms. This can reduce the material cost compared to cadaveric dissection (\cite{george2010review}) and the need for a separate effort of data generation. To this end, we introduce Asynchronous Multi-Body Framework Plus (AMBF+), an open-sourced simulation framework that combines the object interaction capabilities of AMBF with data generation capabilities similar to those of VisionBlender. We also propose a plugin setup described in \autoref{ssec:plugin-design} to accommodate specialized functionalities for different surgical procedures and different data needed for downstream algorithms while keeping the core simulation logic intact. By unifying the surgical training simulation and data generation into one single framework, we attempt to present a synergistic solution to both human training and algorithm development. This framework overcomes the aforementioned limitations and allows surgeons to perform tasks similar to real surgeries while generating/recording relevant information for vision algorithms.

\begin{figure}[htpb]
    \centering
    \includegraphics[width=\textwidth]{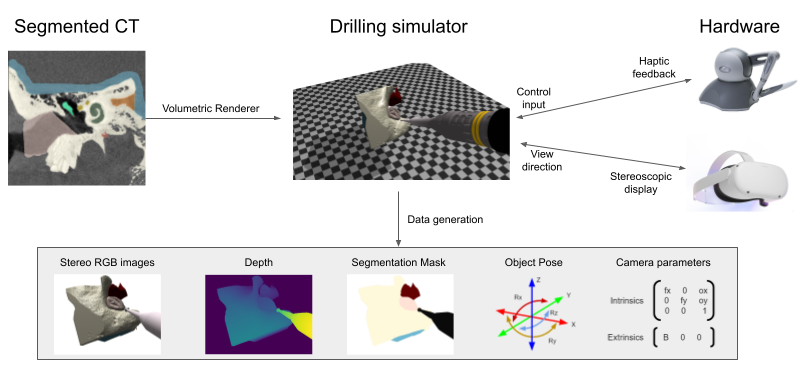}
    \caption[]{Overview of the virtual drilling simulator for skull-base surgery developed using AMBF+. A segmented CT is rendered as a 3D virtual phantom in the simulator. Users send control input and receive haptic feedback with Phantom OMNI\protect\footnotemark[3]. Stereoscopic views are provided with Oculus Rift\protect\footnotemark[4] and the view direction is synchronized with the user's head.}
    \label{fig:overview}
\end{figure}

To demonstrate the functionalities of AMBF+, we have developed a simulator for skull-base surgery similar to \cite{locketz2017anatomy}. Skull-base surgery is a surgical subspecialty in neurosurgery and otolaryngology for surgical intervention of pathology within the base of the skull. During surgery, surgeons inspect the surgical field via surgical microscopes and use surgical drills to remove the bony tissue and gain access to relevant anatomy. In our virtual drilling simulator as shown in \autoref{fig:overview}, users can actively drill a patient model using input devices, while synthetic data is automatically generated. We further demonstrate the utility of this generated data in different computer vision-based tasks.

The key aspects of our development can be summarized as follows:

\begin{itemize}
    \item A virtual surgical simulation framework AMBF+ is provided to enable users to interact with a patient model for surgical practice. The simulation provides both visual and haptic feedback for assistance and realism.
    \item The AMBF+ framework can generate data (RGB stereo images, depth and etc.) in real time while the user is performing virtual surgeries, which can be used for computer vision algorithm development and training.
    \item We propose a plugin design to expand the AMBF+ framework without modification to the core source-code such that functionalities customized for different surgeries can be easily incorporated. 
    \item We demonstrate one use case of AMBF+ -- a virtual drilling simulator for skull-base surgery. We show how the generated data can be potentially used in downstream tasks.
\end{itemize}
\footnotetext[3]{https://www.3dsystems.com/haptics-devices/touch}
\footnotetext[4]{https://www.oculus.com/facebook-horizon}

% \section{Prior Work}

\section{AMBF+ System Overview}
\begin{figure}[htpb]
    \centering
    \includegraphics[width=0.9\linewidth]{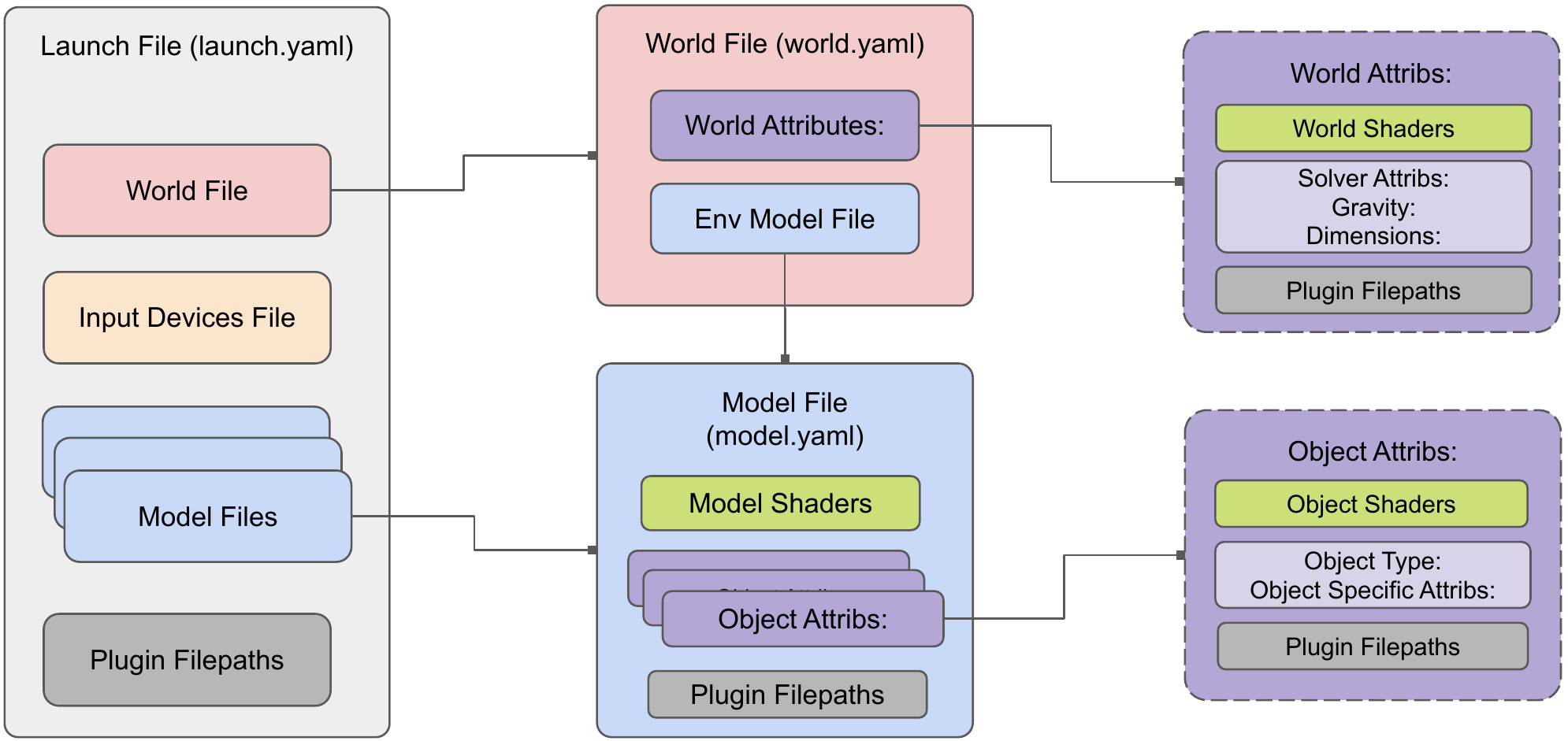}
    \caption{Flow of loading models and world scene objects into AMBF+.}
    \label{fig:ambf_simulator_launch_flow}
\end{figure}

We developed AMBF+ on top of AMBF (\cite{munawar2019real}), which uses libraries such as CHAI3D (\cite{conti2005chai}) and Bullet Physics (\cite{coumans2015bullet}). AMBF+ offers a convenient platform for surgical robot simulations as it allows for the creation of kinematically redundant robots and mechanisms with ease. The framework utilizes an AMBF Description Format (ADF) for defining rigid and soft-bodies, sensors, actuators and world scene objects. The ADF files are modular such that a model (consisting of bodies, joints, sensors and actuators) can be defined in multiple ADF files that can be loaded together. The parenting of bodies, joints, sensors and actuators is also modular which means that a prospective parent or child does not need to exist in the same ADF file. This makes AMBF+ versatile for surgical robot related simulations as surgical robots often have multiple interchangeable tools. The ADF files are of three different types: (1) world files, (2) input devices files (\cite{munawar2021framework}) and (3) model files. A meta-data file, called the \textit{launch file}, is the entry point for AMBF+ and contains the filepaths of a world file, an input devices file and possibly several model files. Fig. \ref{fig:ambf_simulator_launch_flow} describes the flow of loading ADF files into AMBF+.

In AMBF+ we not only extended the rendering pipeline and the ADF specification, but we also implemented a modular plugin handling interface. The extension to the rendering pipeline allowed us to generate and stream colored point-cloud data and segmentation masks. Extension to the ADF specification allowed us to define custom OpenGL shaders for rendering and computation. The plugin handling interface allowed the development of custom applications such as the volumetric-drilling simulation, thus extending its feature set.

\begin{figure}[htpb]
    \centering
    \subfloat[\centering]{{\includegraphics[width=0.3\linewidth]{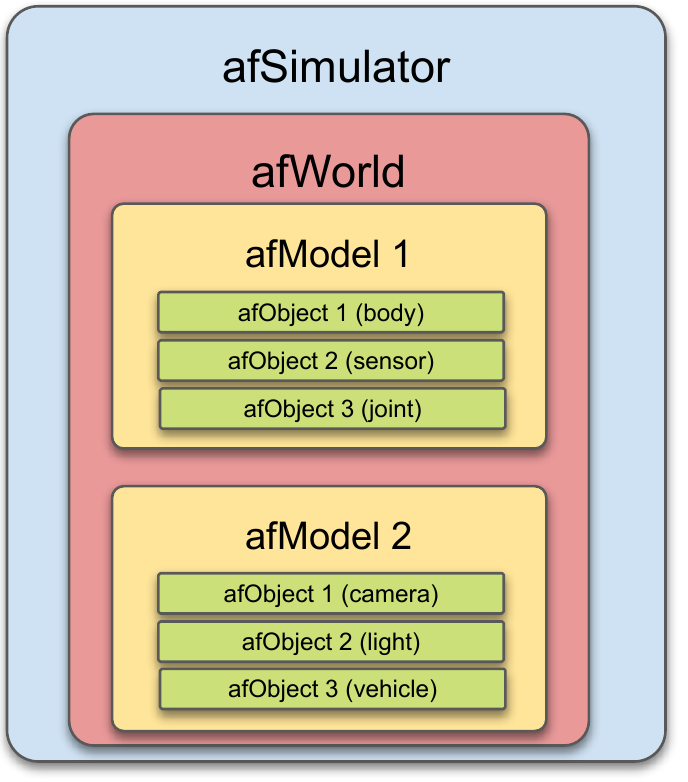} }}%
    \qquad \qquad
    \subfloat[\centering]{{\includegraphics[width=0.3\linewidth]{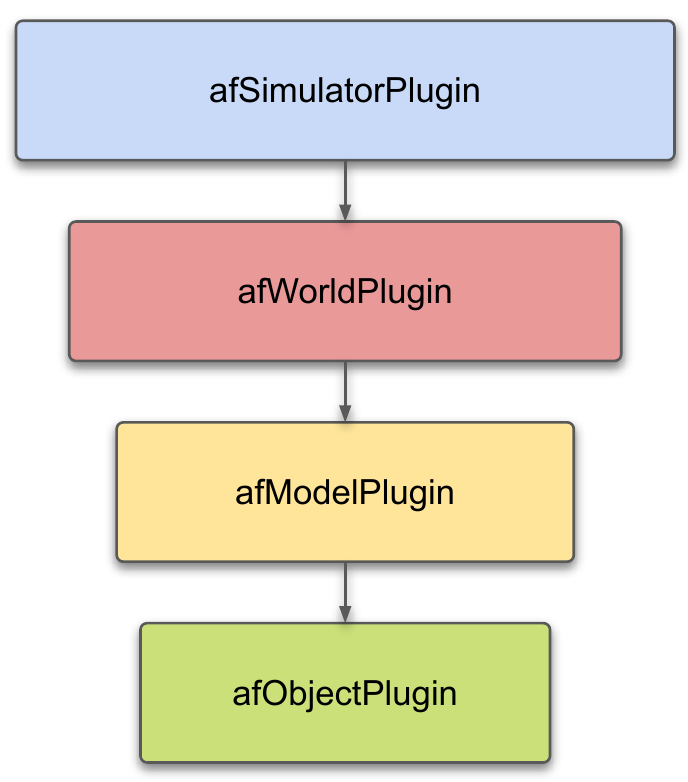} }}%
    \caption{(a) The different computational scopes of an AMBF+ simulation. (b) Corresponding plugins for each computational scope.}%
    \label{fig:ambf_simulaton_scope_and_plugin_hierarchy}%
\end{figure}

\subsection{Plugin Design}
\label{ssec:plugin-design}
\begin{figure}[h]
    \centering
    \includegraphics[width=\textwidth]{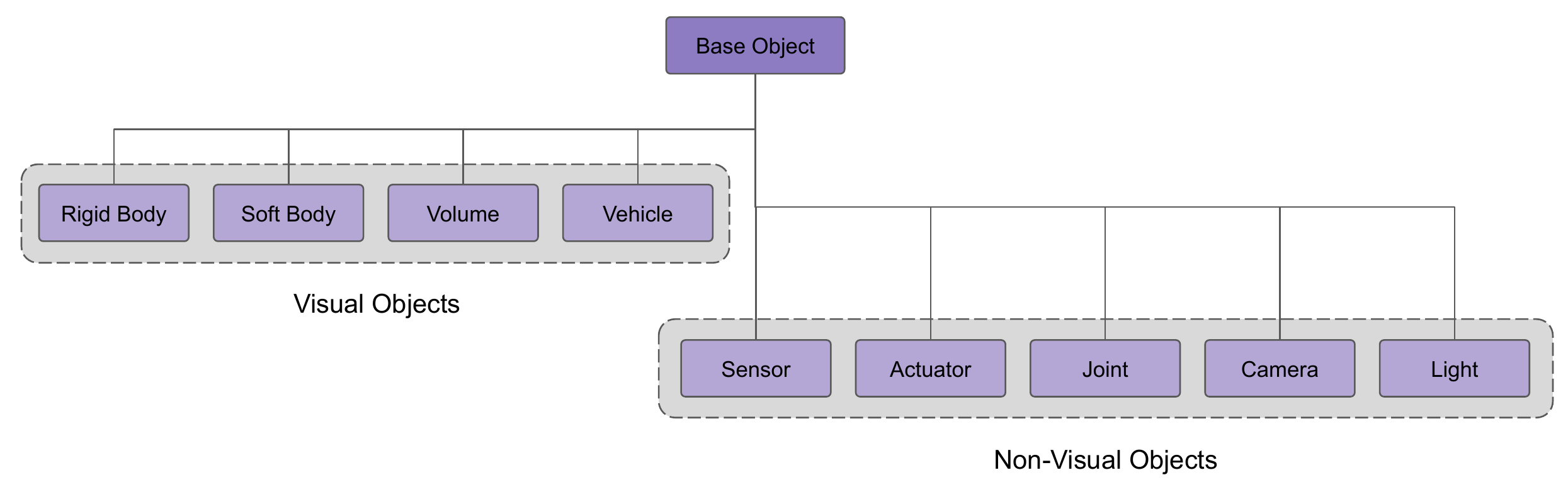}
    \caption{Different types of \textit{afObjects}.}
    \label{fig:ambf_object_types}
\end{figure}

An AMBF+ simulation comprises of different hierarchical computational scopes as shown in Fig. \ref{fig:ambf_simulaton_scope_and_plugin_hierarchy}(a), where each inner block is encapsulated by the outer block to represent a higher scope in hierarchy. A higher scope has access to more resources and control over the simulation components as compared to the lower ones. The highest scope is a single \textit{afSimulator} instance that manages an \textit{afWorld} instance. The \textit{afWorld} may contain instances of \textit{afModels} and each \textit{afModel} may contain several instances of \textit{afObjects}. The term \textit{afObject} refers to simulation entities such as cameras, lights, bodies (rigid and soft), volumes, sensors, and actuators as shown in Fig. \ref{fig:ambf_object_types}. The \textit{afObjects} can be further classified into visual and non-visual objects. Rigid-bodies, soft-bodies, and volumes are types of visual objects.

\begin{figure}%
    \centering
    \includegraphics[width=\linewidth]{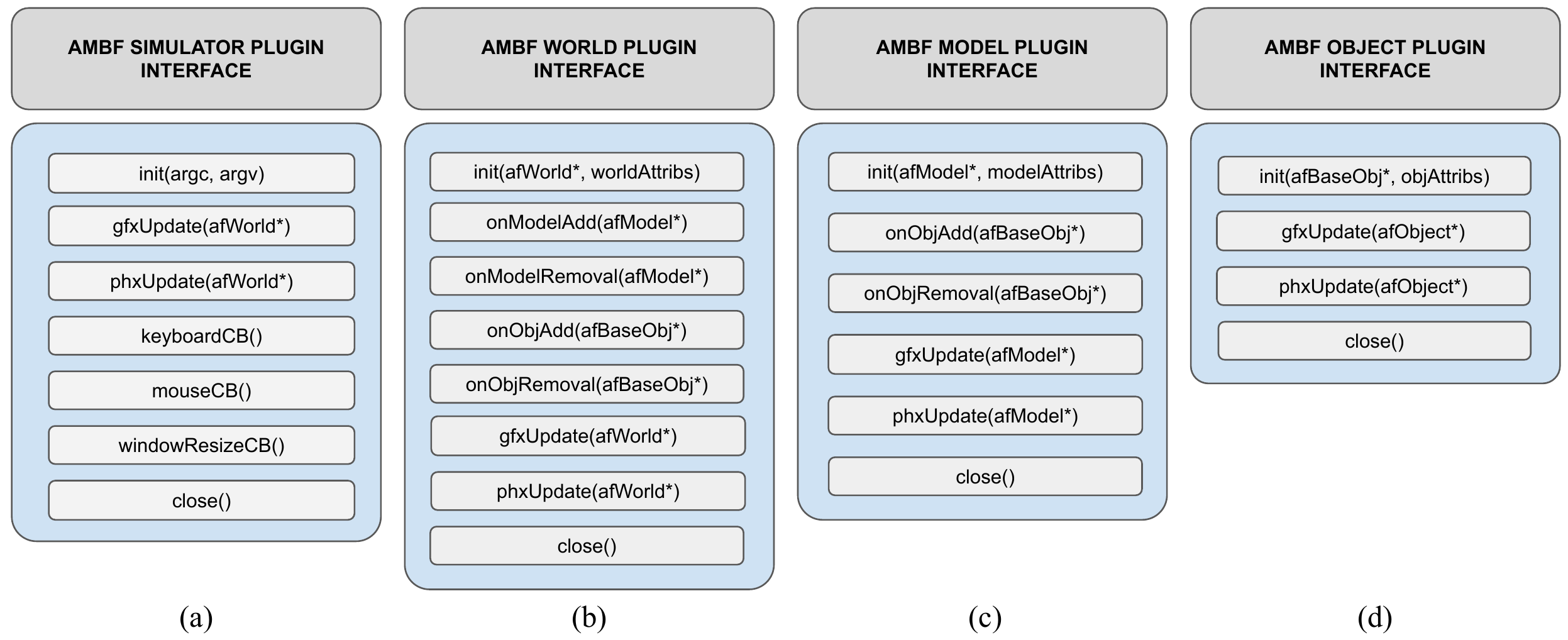}
    \caption{Interfaces provided by the different types of plugins.}%
    \label{fig:ambf_plugin_apis}%
\end{figure}

We developed plugin interfaces for each AMBF+ simulation scope (Fig.~\ref{fig:ambf_simulaton_scope_and_plugin_hierarchy}(b)). Each type of plugin has a slightly different interface, as shown in Fig. \ref{fig:ambf_plugin_apis}, and thus extends the functionality of the scope that it is defined for.

Since the entry to AMBF+ is via the \textit{launch file} (Fig. \ref{fig:ambf_simulator_launch_flow}), we extended the ADF format to incorporate the plugin specification. Plugins can be defined for \textit{afSimulator}, \textit{afWorld} and \textit{afObject} instances by setting them in the appropriate ADF file and/or scope. Our extension to the ADF specification is backward compatible, as older versions of AMBF would simply ignore the \textit{plugins} data-field.

\subsection{Rendering Pipeline}
\label{ssec:render_pipline}
AMBF uses OpenGL\footnotemark[5] for rendering but provides limited access to some of its useful features such as loading custom shaders (vertex and fragment). In AMBF+, we extended AMBF to allow the specification of these shaders via the ADF files. Similar to plugins, shaders can be defined in the world ADF file, the model ADF files, and for individual \textit{afObjects} as shown in Fig. \ref{fig:ambf_simulator_launch_flow}. The shaders are then applied to all afObjects, afObjects in the specific model or only to the specific afObject depending upon the scope that they are specified in. Furthermore, if shaders are defined at multiple scopes, the afObject defined shaders override the model defined shaders and the model defined shaders override the world defined shaders.

\footnotetext[5]{https://www.opengl.org/}
\footnotetext[6]{https://www.khronos.org/opengl/wiki/Depth\_Test}
\footnotetext[7]{https://www.khronos.org/opengl/wiki/Framebuffer\_Object}

% \begin{figure}[htb]%
%     \centering
%     \subfloat[\centering]{{%
%         \includegraphics[width=0.24\linewidth,valign=c]{"img/AMBF Simulator Plugin".pdf}%
%         \vphantom{\includegraphics[width=0.24\linewidth,valign=c]{"img/AMBF World Plugin".pdf}} }}%
%     \subfloat[\centering]{%
%         {\includegraphics[width=0.24\linewidth,valign=c]{"img/AMBF World Plugin".pdf} }}
%     \subfloat[\centering]{{%
%         \includegraphics[width=0.24\linewidth,valign=c]{"img/AMBF Model Plugin".pdf} 
%         \vphantom{\includegraphics[width=0.24\linewidth,valign=c]{"img/AMBF World Plugin".pdf}} }}%
%     \subfloat[\centering]{%
%         {\includegraphics[width=0.24\linewidth,valign=c]{"img/AMBF Object Plugin".pdf} 
%         \vphantom{\includegraphics[width=0.24\linewidth,valign=c]{"img/AMBF World Plugin".pdf}} }}%
%     \caption{Interface provided by the different types of plugins.}%
%     \label{fig:ambf_plugin_apis}%
% \end{figure}

\subsubsection{Point Cloud Generation}
We utilized OpenGL's Z-buffer\footnotemark[6] and Framebuffer objects\footnotemark[7] to generate the point cloud data. Our application utilizes perspective cameras from OpenGL, which induce non-linearity in the generated depth values. The non-linear depth values need to be linearized. This step can be performed on the GPU for faster processing, (in the fragment shader (Alg. \ref{alg:ambf_depth_computation_algorithm})) however the values must also be normalized to avoid truncation by the fragment shader's output. Finally, on the CPU, these normalized values are rescaled to get the correct point could. The maximum dimensions ($\vec{MD}$) of the view frustum are used for both the normalization and scaling. The vector $\vec{MD}$ is calculated as:

% We utilized OpenGL's Z-buffer\footnote{\url{https://www.khronos.org/opengl/wiki/Depth_Test}} and Framebuffer objects\footnote{\url{https://www.khronos.org/opengl/wiki/Framebuffer_Object}} to generate the point cloud data. Our application utilizes perspective cameras from OpenGL, which induce non-linearity in the generated depth values. The non-linear depth values need to be linearized and scaled appropriately to get an accurate point cloud. For faster processing, the linearization can be performed on the GPU (in the fragment shader (Alg. \ref{alg:ambf_depth_computation_algorithm})), but the scaling needs to be performed on the CPU as the fragment shader's output must be normalized. The scaling depends upon the maximum dimensions ($\vec{MD}$) of the view frustum which is calculated as:

% Due to the non-uniform shape of the view frustum for perspective cameras, the point cloud data values along the horizontal, vertical, and depth axes are of different scale. We call this scale as the maximum dimensions ($\vec{MD}$) and it is required for our depth computation algorithm Alg. \ref{alg:ambf_depth_computation_algorithm}. $\vec{MD}$ is calculated as follows:

\begin{equation}
    MD_x = 2.0 \times f \times tan(fva / 2); \quad MD_y = MD_x \times AR; \quad MD_z = f - n
\end{equation}

Where $n$, $f$, $fva$ and $AR$ correspond to the simulated camera's near plane, far plane, field view angle and aspect ratio, respectively.

\begin{algorithm}[htpb]
\small
\caption{Depth Computation Shader}
\begin{algorithmic}[1]
\State $F_x, F_y \gets FragCoord.xy$ \algorithmiccomment{Input to the Shader}
\State $B_0,B_1,B_2,B_3 \gets Texture2D(F_x, F_y)$ \algorithmiccomment{Depth packed in 4 one-byte channels}
\State $Z \gets (B_3 \ll 24) \lor (B_2 \ll 16) \lor (B_1 \ll 8) \lor B_0$ \algorithmiccomment{Bit Shifting and Logical OR}
\State $F_z \gets Z / 2^{24}$
\State $P_{norm} \gets [F_x, F_y, F_z, 1.0] $
\State $P_{clip} \gets (ProjectionMatrix)^{-1} P_{norm}$ \algorithmiccomment{Projection Matrix of Simulated Camera}
\State $P_{cam} \gets P_{clip} / P_{clip}.w$
\State $N_{xy} \gets (P_{cam}.xy + \vec{MD}_{xy} / 2.0) / \vec{MD}_{xy}$
\State $N_z \gets (P_{cam}.z - n) / (f - n)$
\State $FragOutput \gets [N_x, N_y, N_z, 1.0]$
\end{algorithmic}
\label{alg:ambf_depth_computation_algorithm}
\end{algorithm}

% Finally, on the client (CPU) side, the normalized Framebuffer (output of the Fragment shader) is re-scaled and offset to get an accurate point cloud with respect to the camera frame.

\subsubsection{Segmentation Mask Generation}
\begin{figure}[htpb]
    \centering
    \subfloat[\centering]{{\includegraphics[width=0.38\textwidth,valign=c]{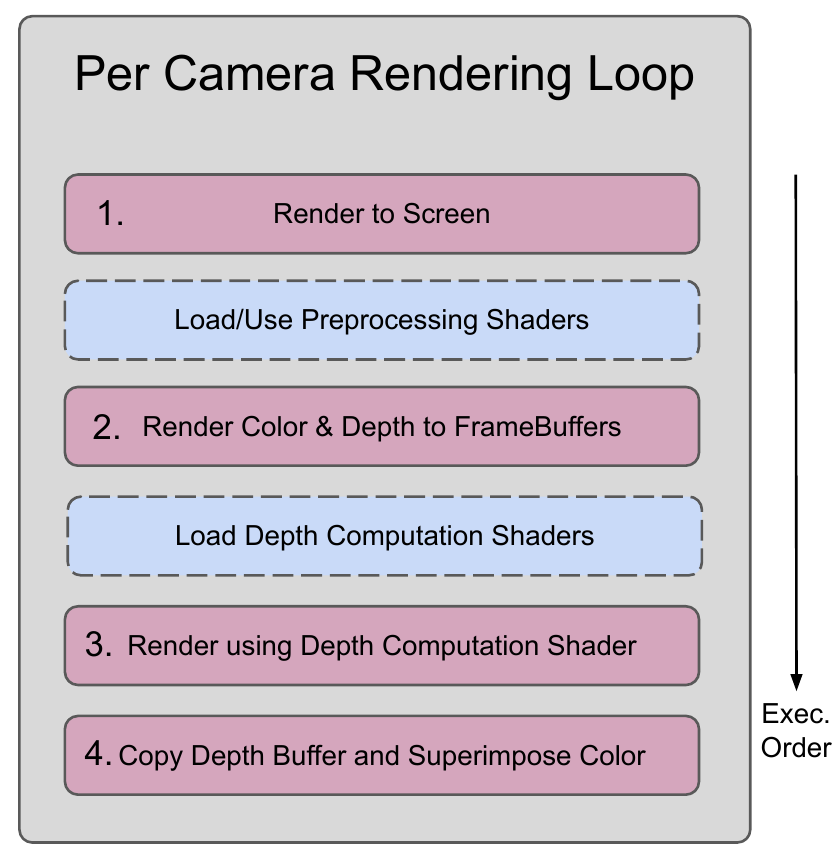} }}%
    \subfloat[\centering]{{\includegraphics[width=0.6\linewidth,valign=c]{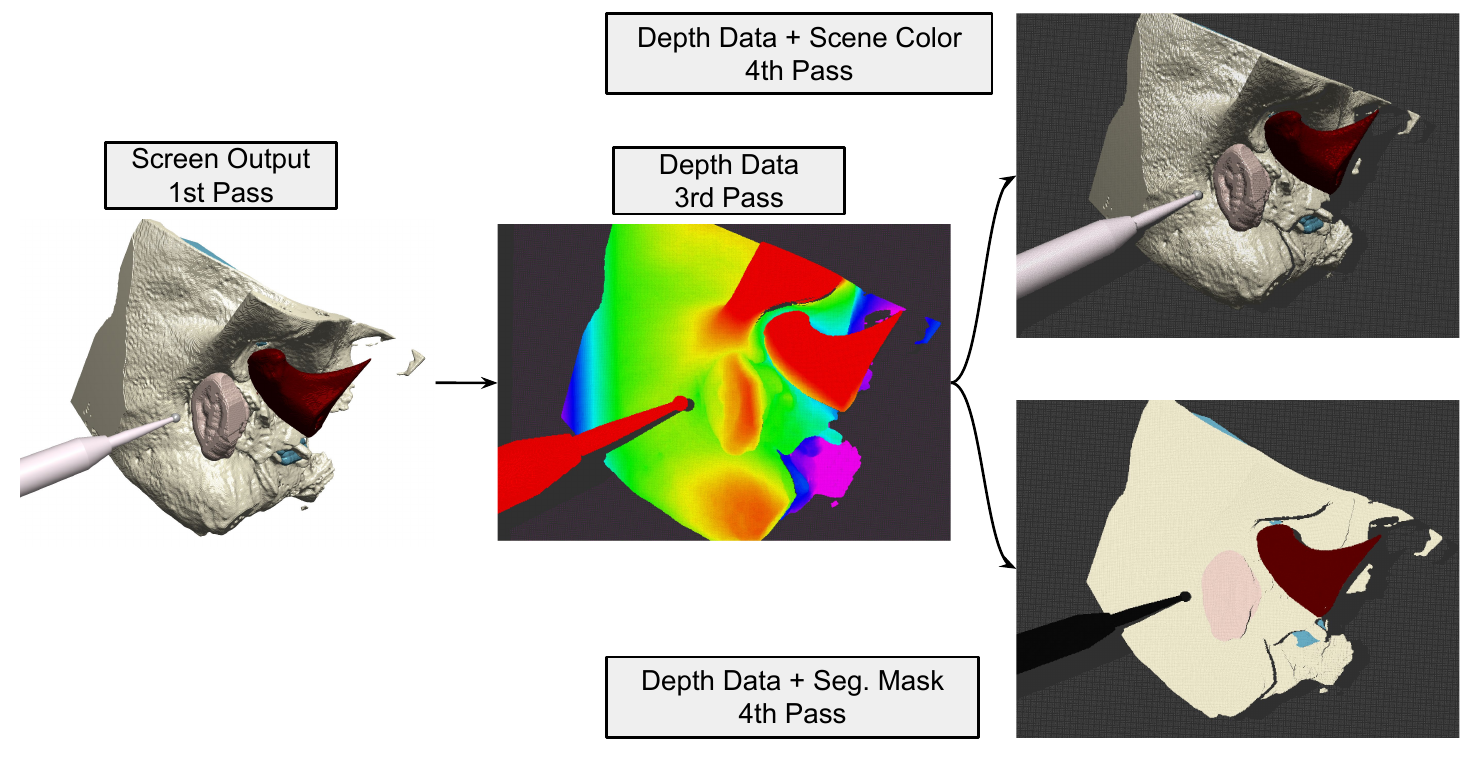} 
    \vphantom{\includegraphics[width=0.38\textwidth,valign=c]{"img/AMBF_Rendering_Pipeline".pdf}}}}
    \caption{(a) Different passes in the rendering pipeline to generate a segmented point cloud. (b) Augmentation of the segmentation mask on the point cloud to generate labeled data.}%
    \label{fig:ambf_rendering_pipeline}%
\end{figure}

AMBF+'s rendering pipeline comprises of four passes which are shown in Fig \ref{fig:ambf_rendering_pipeline}(a). Three passes are implemented on the GPU and the fourth one on the CPU. The first pass generates a 2 dimensional color image that is output to the screen. The second pass renders to the color and depth Framebuffers. The values in the depth-buffer are non-linear at this point. Prior to the third pass, the depth computation shaders are loaded. These shaders are responsible for linearizing and normalizing the depth values (Alg. \ref{alg:ambf_depth_computation_algorithm}) on the third rendering pass. The fourth pass re-scales the normalized depth values and computes camera space transformation of the point cloud data. During this pass, the color Framebuffer generated from the second pass is also superimposed on the point cloud to generate a colored point cloud as shown in Fig. \ref{fig:ambf_rendering_pipeline}(b).

We utilize the custom shader specification, via the ADF files, to load special \textit{pre-processing shaders} for visual objects prior to the second rendering pass. The sole purpose of these shaders is to assign a specific color to the visual object while ignoring the lighting calculations. The result is a segmentation mask image that can be superimposed on the point cloud in the fourth pass as shown in Fig. \ref{fig:ambf_rendering_pipeline}(b).

\subsection{Data Streaming and Recording}
\label{ssec:data_streaming}
We use the Robot Operating System (ROS)\footnotemark[8] interface due to its wide usage in the surgical robotics community. We stream out the data generated from AMBF+ as ROS topics. % and record the items listed below in an \textit{HDF5} file. 
Such design allows for replay of surgical operation, easy retrieval and efficient storage. \footnotetext[8]{{https://www.ros.org/}}

\begin{itemize}
    \item Stereo images: Stereo images (left and right) are published via a ROS image topic by AMBF+ at a user specified frequency. The left and right stereo images are produced by two different cameras initialized in AMBF+. By manipulating the OpenGL parameters of the two cameras, \textit{location} and \textit{look at}, the stereo view can be set up in a highly customizable fashion. \textit{Location} is used to establish the stereo baseline and \textit{look at} is used to align the principle axes of the left and right cameras.
    \item Depth: A depth point cloud is created as described in \textit{Section 2.3.1}. %, is published as a ROS PointCloud2 message. This message is streamed through the left stereo microscope camera topic. 
    \item Segmentation mask: The ground truth segmentation mask is created as described in \textit{Section 2.3.2}. % is published as both ROS PointCloud2 message and as an image.
    \item Object/camera pose: Object/camera pose contains the xyz position and rotation (either quaternion or Euler angles) with respect to the world frame. Different objects and cameras may be added to the AMBF+ scene and can be tracked and controlled by AMBF+ and its Python client.
    \item Camera parameters: The intrinsic parameters of cameras in AMBF+ are established in OpenGL fashion using the vertical field view angle ($fva$) which describes the perspective frustum. The intrinsic matrix ($I$)  can be calculated through the following equations:

\vspace{-0.5cm}
\begin{equation}
\ I = \begin{bmatrix}
f_x & 0 & c_x\\
0 & f_y & c_y\\
0 & 0 & 1\\
\end{bmatrix};\quad
f_x = f_y = \frac{H}{2\times tan\left(\frac{fva}{2}\right)};\quad
 c_x = W/2; \quad
 c_y = H/2
\end{equation}

Where $f_x$ and $f_y$ are the focal length in pixels in the x and y directions. $W$ and $H$ are the width and height of the image produced in pixels. $c_x$ and $c_y$ describe the x and y position of the camera principal point. Extrinsics information is obtained from the relative transformation between the two cameras.

\end{itemize}

\subsection{Stereo Display and VR Support}
Depth perception is a requirement for a simulation environment targeting surgeries. We generate a pair of stereoscopic images that are then displayed on a Virtual Reality (VR) head-mounted display (HMD). We also support the orientation control of the virtual cameras based on the HMD's angular movements as a means of looking around in the simulation scene. We rely on the OpenHMD\footnotemark[9] package for HMD support and tested our implementation using an Occulus Rift VR device.
\footnotetext[9]{https://github.com/OpenHMD/OpenHMD}

\subsection{Haptic Feedback}
AMBF+ utilizes CHAI3D's (\cite{conti2005chai}) finger proxy collision algorithm (\cite{ruspini2001haptic}) to provide haptic feedback of interaction with surface meshes and volumes. Tool-cursors, where each tool-cursor encapsulates a pair of proxy/goal spheres, are used for feedback calculation. The movement of the tool-cursor moves the encapsulated proxy/goal spheres such that the proxy sphere is stopped at the boundary of the surface or volume while the goal sphere coincides with the tool-cursor's pose, and is thus allowed to penetrate inside. The distance between the proxy and the goal spheres ($\delta p_i$) for each tool-cursor can then be used to generate a proportional force feedback.

\subsection{Use-case of AMBF+ -- Virtual Drilling Simulator}
To demonstrate the flexibility of AMBF+, a drilling simulator is developed that leverages the aforementioned functionalities. The scheme of the simulator is summarized in \autoref{fig:overview}. We customize the following functionalities using the plugin design: 1) customized data representation from a pre-operative CT scan, which will be rendered at frame-rate, 2) virtual drilling with input from a haptic device (Phantom OMNI or Oculus Rift in our setup). A video demonstration of our simulation scene is provided in the supplementary material.

\subsubsection{Virtual Model Representation}
The virtual patient model is obtained from a CT scan and is patient specific. We use an off-the-shelf CT segmentation pipeline (\cite{ding2021automated}) to generate the corresponding segmented volume for each anatomical structure in 3D. The segmented volume is used for three purposes: 1) to texture each anatomy with its corresponding anatomical color, 2) to differentiate skull and other critical anatomies for visual warnings, and 3) to generate 2D segmentation masks.

We use an array of 2-dimensional images (in PNG or JPEG format) for texture-based volumetric rendering (\cite{kruger2003acceleration}) and drilling. These images combined represent the 3D volume, with each color denoting a different anatomy. We provide utility scripts that can convert the data from the NRRD format to an array of images. This allows the segmented volumetric data to be imported for use within the virtual drilling simulator. 
% In our experiment, our volumetric data is of dimension $512\times512\times512$ ($width \times height \times image\, slices$). 

\subsubsection{Virtual Drilling and Haptic Feedback}
\begin{figure}[htpb]
    \centering
    \subfloat[\centering]{{\includegraphics[width=0.4\textwidth]{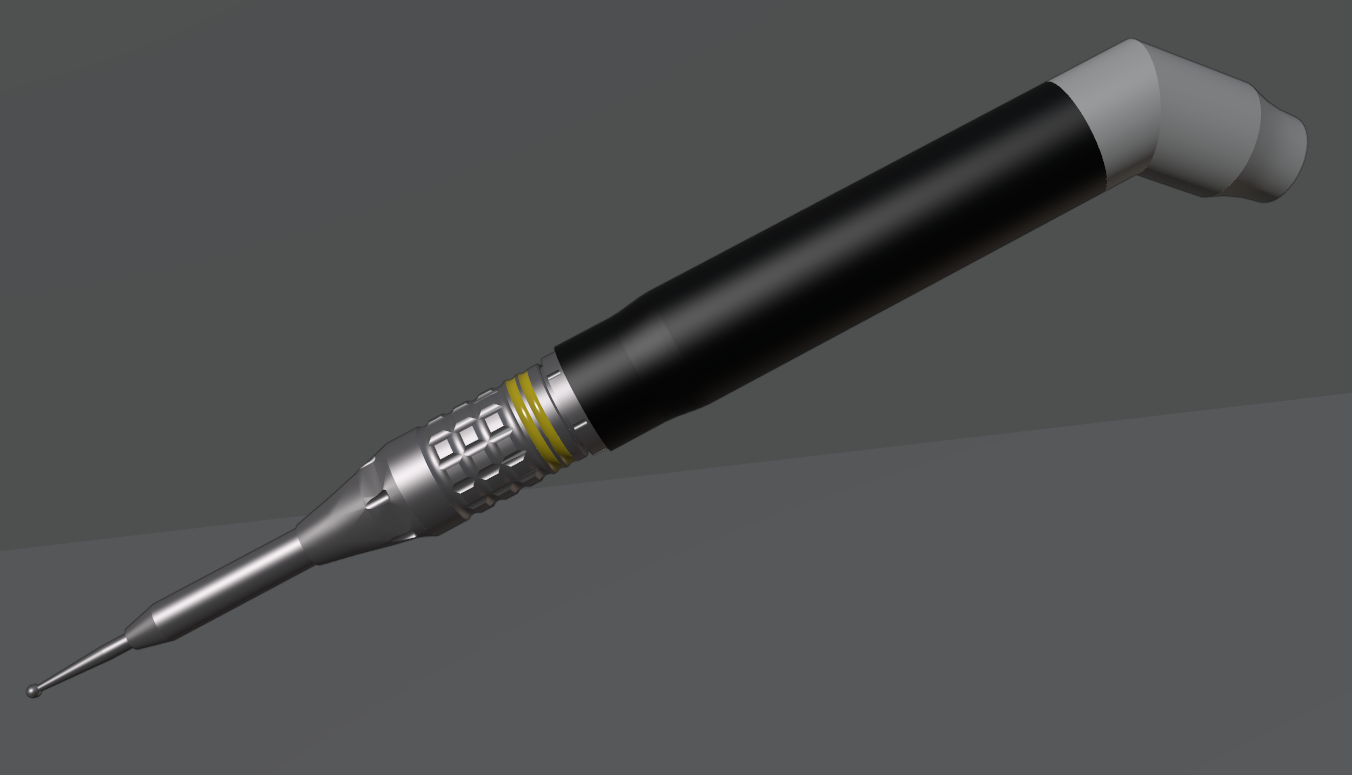} }}%
    \subfloat[\centering]{{\includegraphics[width=0.409\linewidth]{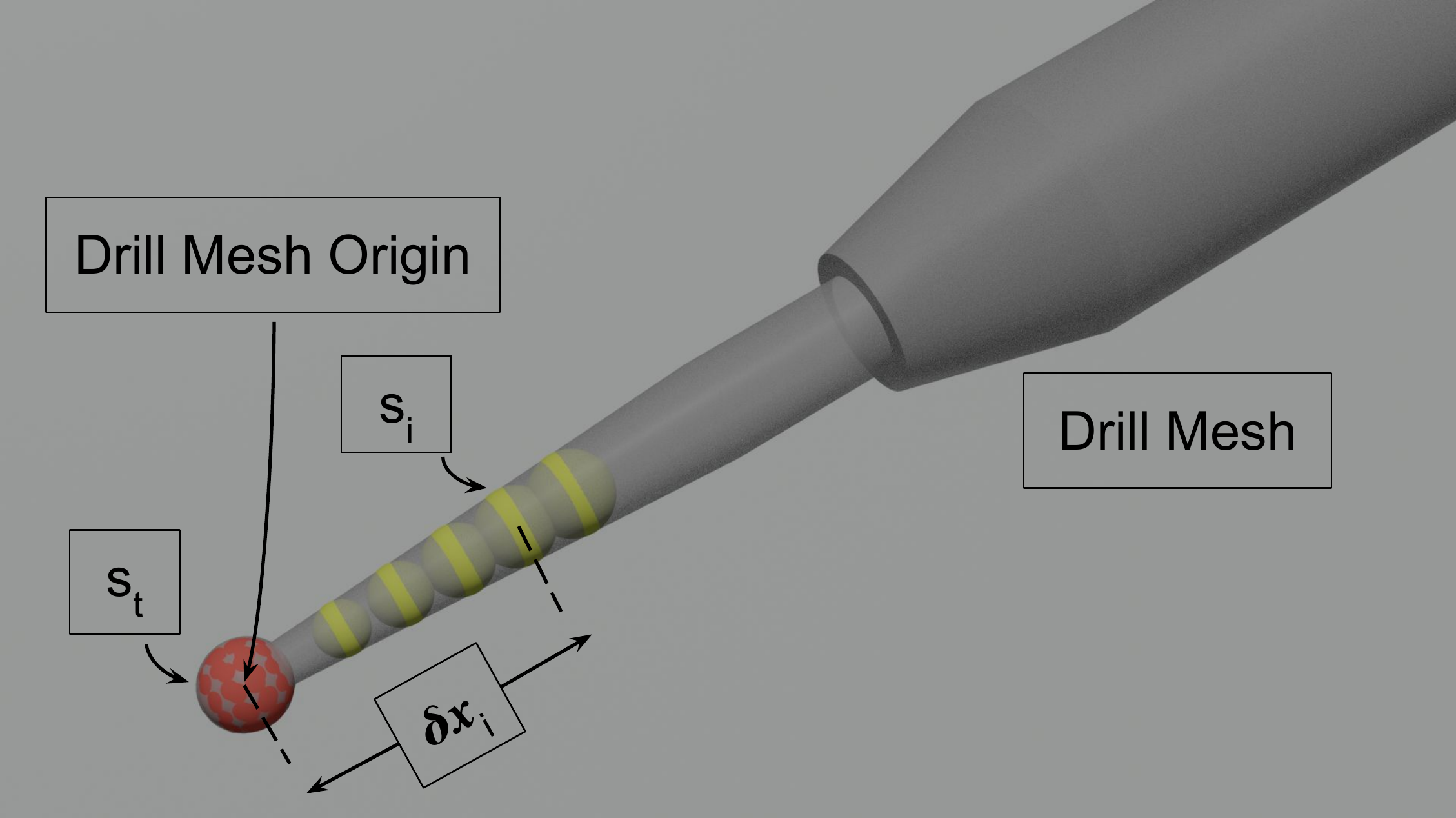} }}
    \caption{(a) Mesh model of the simulated drill. (b) Collision approximation of the simulated drill with tip and shaft tool-cursors. The origin of the drill mesh coincides with the center of $s_t$.}%
    \label{fig:shaft-collision-approx}%
\end{figure}

Our drilling simulator allows the users to drill the volumetric data (anatomy) with a haptic device which is represented as a simulated drill, as shown in Fig. \ref{fig:shaft-collision-approx}. The CHAI3D (\cite{conti2005chai}) library is used for haptic feedback logic. For a more accurate collision response, the simulated drill is approximated by one tool-cursor ($s_t$) at the tip and multiple tool-cursors ($s_i,\, i \in [0,1,2 ...]$) placed along the shaft's axis. The tip's tool-cursor is distinguished from the shaft tool-cursors, since only the tip of the simulated drill can remove bone tissue. When the tip tool-cursor interacts with the anatomy, the set of colliding voxels are retrieved and their intensity is set to 0. This causes the volume to be ``drilled" on the next rendering update. On the other hand, if $s_i$ interacts with the anatomy, the drill is prevented from drilling and penetrating further. Alg. \ref{alg:tip_and_shaft_collision} summarizes our approach.

\begin{algorithm}[htpb]
\small
\caption{Drill Shaft Collision Algorithm}
\begin{algorithmic}[1]
\State $T_D :=$ Simulated drill's pose
\State $T_s := I_{4\times4}$  \algorithmiccomment{$I:= $Identity Matrix}
\State $\vec{e}_{max} \gets \vec{0}$
\State $s_t \gets$ Tip tool-cursor
\State $S \gets $ Shaft tool-cursors [$s_n$, $...$, $s_1$, $s_0$]
\State $\delta \vec{x_i} :=$ Fixed aligned distance between shaft tool-cursor $s_i$ and $s_t$ \algorithmiccomment{$i \in [0,1,2,..]$}
\ForEach {$s_i \in S$}
\State  $\delta \vec{p}_i \gets compute\_error(s_i)$ \algorithmiccomment{Error between proxy and goal}
\If {$||\delta \vec{p}_{i}|| > ||e_{max}||$}
\State $[\vec{P_s},\,  R_s] \gets Pose(s_i.proxy)$ \algorithmiccomment{$\vec{P_s} :=$ Position of $s_i$ proxy}
% \State $R_s \gets Rotation(s_i.proxy)$ \algorithmiccomment{$R_i :=$ Rotation Matrix of $s_i$ proxy}
\State $T_s \gets \vec{P_s} + R_s \delta \vec{x_i}$ \algorithmiccomment{$R_s :=$ Rotation Matrix of $s_i$ proxy}
\State $s_{max} \gets s_{i}$ 
\EndIf
\State $\vec{e}_{max} \gets $ max$(\vec{e}_{max}, \delta \vec{p}_i)$
\EndFor
\If {$||\vec{e}_{max}|| > \epsilon \wedge s_{max} \neq s_t$} \algorithmiccomment{$\epsilon :=$ A small value for numerical stability}
\State $T_D \gets T_s$ 
\State $\vec{F} \gets control\_law(\vec{e}_{max})$ \algorithmiccomment{$\vec{F} :=$ Force feedback}
\Else
\State $T_D \gets Pose(s_t.proxy)$
\State $remove\_colliding\_voxels()$ \algorithmiccomment{Set the intensity of colliding voxels to zero}
\State $\vec{F} \gets control\_law(compute\_error(s_t))$ 
\EndIf
\end{algorithmic}
\label{alg:tip_and_shaft_collision}
\end{algorithm}

\section{Evaluation and Discussion}
In this section, we present evaluation on the data quality for algorithm development. We will evaluate the surgical training components with subsequent user studies in our future work.
\subsection{Geometric-based Computer Vision Algorithm - Anatomy Tracking}
\label{ssec:orbslam}
Vision-based pose tracking is a task where input images are processed to infer the pose information of objects. It recently has found applications in Augmented Reality (AR) and can potentially improve the safety of skull-base surgeries when used as a navigation system (\cite{mezger2013navigation}). We demonstrate how the data generated from the simulator can be used for evaluating vision-based tracking algorithms. In the experiment, we use the state-of-the-art ORB SLAM V3 algorithm by \cite{campos2021orb}. The algorithm extracts ORB features (\cite{rublee2011orb}) from stereo images and uses the matched features to compute the camera pose relative to the patient's anatomy based on geometric consistency. We refer interested readers to the paper for more details. A video demonstration of ORB SLAM V3 running on our data can be found in the supplementary material.

\begin{figure}
    \centering
    \subfloat[]{\includegraphics[height=0.45\textwidth]{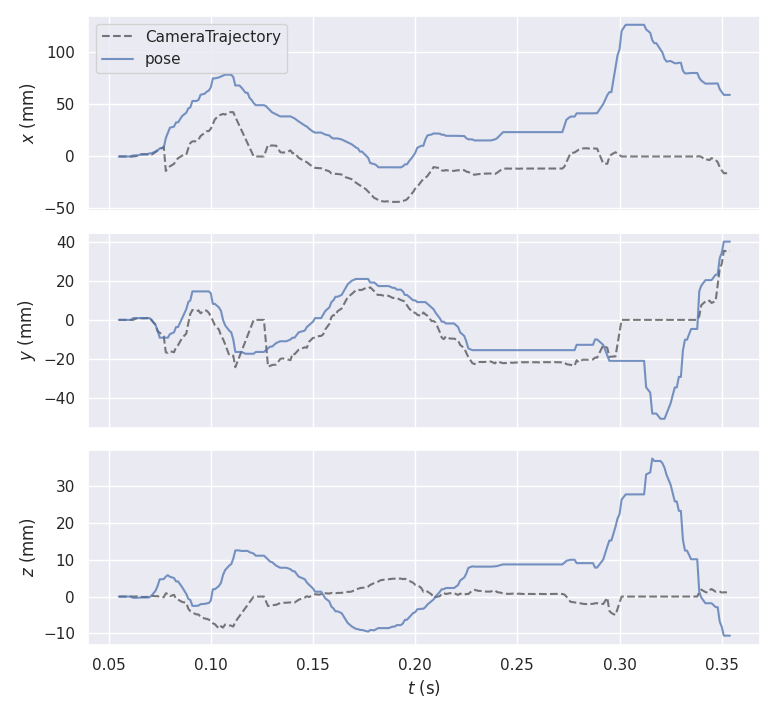}}
    \hspace{0.4em}
    \subfloat[]{\includegraphics[height=0.45\textwidth]{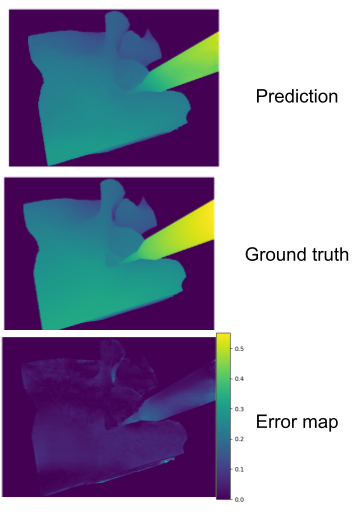}}
    
    \caption{(a) Plot of estimated XYZ trajectory of anatomy pose tracking from ORB SLAM V3. (b) Visualization of depth prediction from STTR.}
    \label{fig:result}
\end{figure}

Since we have ground truth pose information available (\textit{c.f.} \autoref{ssec:data_streaming}), we can directly evaluate the accuracy of the algorithm outputs. The baseline of the generated stereo images is set to 65\,mm following stereo microscope design. We evaluate the algorithm in two types of settings: 1) static tool with camera movement, and 2) static camera with tool movement and anatomy modification. In each setting, 500 frames were recorded and used for evaluation. These sets of experiments can evaluate the accuracy of ORB SLAM V3 when the camera moves and show if the tool movement contaminates the pose tracking. We summarize the result in \autoref{tab:tracking_result}. While the ORB SLAM V3 algorithm performs very robustly when only the drill moves, the tracking accuracy degrades with camera motion. We plot the translation error in \autoref{fig:result}. The tracking error with camera movement is on the order of 40\,mm for translation and around 8 degrees for rotation, which is still impractical for clinical use directly (0.5 mm as in \cite{schneider2020evolution}). We also observed frequent tracking failures when camera movement is large. This suggests future research opportunities of improving tracking algorithms with microscopic data. The analysis with simulation generated data can serve as a complement to image quality based analysis in \cite{long2021dssr} using real images.

\begin{table}[htpb]
\centering
\caption{Quantitative result (mean and standard deviation of L1 error) of ORB SLAM V3 applied on the synthetic stereo microscopic data generated by the drilling simulator.}
\label{tab:tracking_result}
\resizebox{0.7\textwidth}{!}{%
\begin{tabular}{c|c|c|}
 & \textbf{Translation Error (mm)} & \textbf{Rotation Error (deg)} \\  \hline
Moving camera & 40.97 $\pm$ 22.40 & 8.44 $\pm$ 3.07 \\
Moving drill & 8.1E-1 $\pm$ 9.1E-1 & 3.2E-3 $\pm$ 3.6E-3
\end{tabular}%
}
\end{table}

\subsection{Deep Learning-based Computer Vision Algorithm - Depth Estimation}
We demonstrate that data generated from our simulator can also be used for deep learning algorithms. We conduct our experiment on the stereo depth estimation task, where stereo images are used to densely estimate the depth of the scene. The moving camera sequence is used as training data and the moving drill sequence is used as testing data (\textit{cf.} \autoref{ssec:orbslam}). In the training data, the camera observes the unmodified anatomy at different locations. In the testing data, the virtual drill actively changes the anatomy with a fixed camera location. We train the stereo depth network STTR (\cite{li2021revisiting}) for depth estimation from scratch with AdamW optimizer for 5 epochs with learning rate 1E-4. The testing L1 error of depth is 1.98\,mm. The qualitative result is shown in \autoref{fig:result}. The trained depth network generalizes to changes of anatomy shape even if it is only trained with constant anatomical shape. The fully controlled simulation environment makes such analysis easy.

We note that our evaluation only demonstrates weak generalization, \textit{i.e.} we train and evaluate the algorithm both in synthetic settings. The performance on real images of the trained network can be limited due to the large domain gap between simulated and real images. 

\section{Conclusion and Future Work}
In this work, we report our progress towards a cost-effective and synergistic framework, AMBF+, unifying surgical training and data generation. It allows users to practice surgical procedures and generates relevant data in real-time simultaneously. The framework is designed to be modular and can be extended for different surgeries and data. We have demonstrated one use case where a virtual skull-base surgery is simulated. We have presented evaluation on data quality from the simulator and use cases of the data in different downstream computer vision tasks. Our future work includes incorporating virtual fixture testing in the simulation environment (\cite{li2020hybrid}, \cite{li2020anatomical}) and a formal user study with surgeons and residents to evaluate the efficacy of our simulator in surgical training. We also recognize the limitation of visual realism and plan to explore learning-based style transfer (\emph{e.g.}, \cite{rivoir2021long}) to mitigate the problem.
% more advanced physics simulation integrating fluid dynamics such that irrigation/suction tools and blood can be integrated 

\section*{Acknowledgement}
This work was supported by: 1) an agreement LCSR and MRC, 2) Cooperative Control Robotics and Computer Vision: Development of Semi-Autonomous Temporal Bone and Skull Base Surgery K08DC019708, 3) Intuitive research grant, 4) a research contract from Galen Robotics, 5) Johns Hopkins University internal funds.

\section*{Disclosures}
Russel H. Taylor is a paid consultant to Galen Robotics and has an equity interest in that company. These arrangements have been reviewed and approved by JHU in accordance with its conflict of interest policy.

\bibliographystyle{tfcse}
\bibliography{interactcsesample}

\end{document}